\documentclass[letterpaper, 10 pt, conference]{ieeeconf}  

\IEEEoverridecommandlockouts                              

\usepackage[hyphens]{url}
\usepackage{hyperref}
\hypersetup{breaklinks=true}

\usepackage[euler]{textgreek}
\usepackage{cite}
\usepackage{xspace}
\usepackage{graphicx}
\usepackage{textcomp}
\usepackage{subfigure}
\usepackage{balance}
\usepackage{multicol}
 \usepackage{amsmath} 
 \usepackage{mathtools}
\usepackage{mathptmx}  
\usepackage{algorithmic}
\usepackage{lipsum}

\usepackage{etoolbox}
\usepackage[linesnumbered,ruled,vlined]{algorithm2e}
\DontPrintSemicolon

\title{\LARGE \bf I can attend a meeting too! Towards a human-like telepresence avatar robot to attend meeting on your behalf
}
\author{Hrishav Bakul Barua, Chayan Sarkar, Achanna Anil Kumar, Arpan Pal, and Balamuralidhar P
\thanks{Hrishav Bakul Barua, Chayan Sarkar, Achanna Anil Kumar, Arpan Pal, and Balamuralidhar P are with TCS Research \& Innovation, India
        {\tt\small \{hrishav.barua, sarkar.chayan, achannaanil.kumar, arpan.pal, balamurali.p\}@tcs.com}}%
}
\begin{document}

\maketitle
\thispagestyle{empty}
\pagestyle{empty}

\begin{abstract}
Telepresence robots are used in various forms in various use-cases that helps to avoid physical human presence at the scene of action. In this work, we focus on a telepresence robot that can be used to attend a meeting remotely with a group of people. Unlike a one-to-one meeting, participants in a group meeting can be located at a different part of the room, especially in an informal setup. As a result, all of them may not be at the viewing angle of the robot, a.k.a. the remote participant. In such a case, to provide a better meeting experience, the robot should localize the speaker and bring the speaker at the center of the viewing angle. Though sound source localization can easily be done using a microphone-array, bringing the speaker or set of speakers at the viewing angle is not a trivial task. First of all, the robot should react only to a human voice, but not to the random noises. Secondly, if there are multiple speakers, to whom the robot should face or should it rotate continuously with every new speaker? Lastly, most robotic platforms are resource-constrained and to achieve a real-time response, i.e., avoiding network delay, all the algorithms should be implemented within the robot itself. This article presents a study and implementation of an attention shifting scheme in a telepresence meeting scenario which best suits the needs and expectations of the collocated and remote attendees. We define a policy to decide when a robot should rotate and how much based on real-time speaker localization. Using user satisfaction study, we show the efficacy and usability of our system in the meeting scenario. Moreover, our system can be easily adapted to other scenarios where multiple people are located.
\end{abstract}


\section{Introduction}
Robots are becoming an integral part of our life and human-robot interaction (HRI) has become an indispensable part of today's robots. The most convincing factor behind this trend is a human's constant urge to interact with machines in a more human-like fashion. Goodrich \textit{et al.}~\cite{goodrich2008survey} presents a comprehensive survey on the role of HRI and its various aspects. Some of the areas where HRI can influence the overall success of the task are search and rescue, assistive robotics, military and police, edutainment, space explorations, home, industry, etc. Interactions can be successful only if a human understands what a robot is up to and a robot understands what a human is indicating or saying. It is known that a robot actually cannot perceive as a human, and it is not expected from them either. But, people at least expect that a robot mimics human behavior and action.

Robot as an avatar is not a science-fiction anymore, but a reality and HRI has a higher degree of importance for an avatar robot. Multi-party meetings, be it in industry, academia or government sector, is a common phenomenon nowadays, which often involves traveling of one/more people over a large distance. However, an avatar robot, if equipped with the right set of functionality, can be used to attend a meeting without physically being present at the location. Such a telepresence setup is beneficial compared to the traditional video conferencing systems as it not only provides a 360\textdegree\, view of meeting proceedings, it facilitates an informal meeting setup and mobility support, which is often required if the meeting activities are not limited to one display screen. A telepresence robot can turn to any person of interest, exchange eye-gazes, record talks, identify speakers through audio-visual multi-modal systems and so on. 

In this work, we aim at defining and designing an HRI scheme for a telepresence meeting scenario where a robot replaces a meeting attendee and behaves like the absent attendee as best as possible without explicit control from a human being. Although some work has been done in this area, less focus has been given on the attention part of the interaction. Considering the fact that giving attention to individual speakers in a meeting and retaining focus are areas of open research from HRI perspective~\cite{stoll2018wait, neustaedter2016beam}, we choose this problem. Existing studies~\cite{rae2017robotic} on various telepresence robots suggest that research and designing should be focused on improving the viewing range of a robot so that the remote attendee can see the meeting proceedings in a way s/he would have seen if attended physically. This will also facilitate efficient face to face communication and interaction and adhere to better user experience.

\begin{figure}
	\centering
	\includegraphics[width=\linewidth]{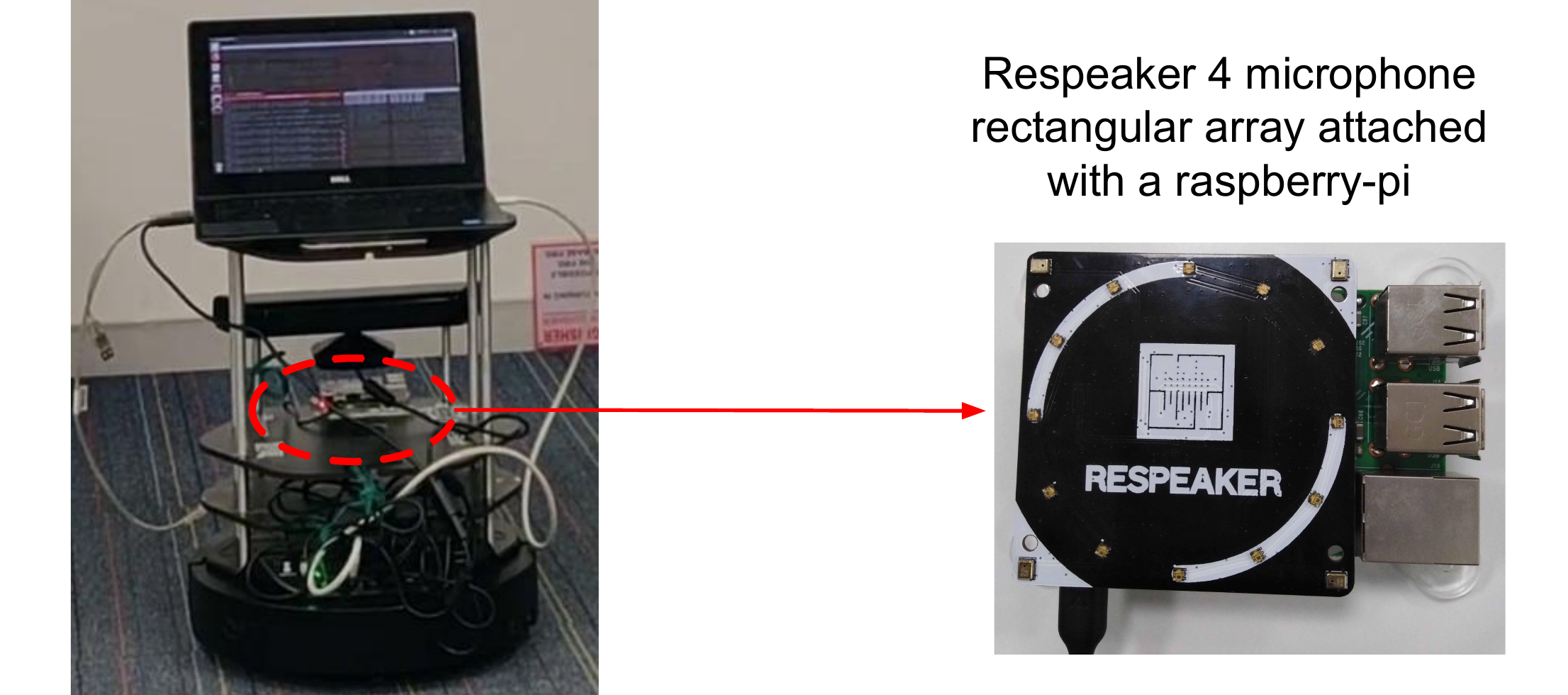}
	\caption{The hardware setup used in our experiments.}
	\label{fig:turtlebot}
\end{figure}

\textbf{Contributions:} We deal with the given problem scenario using sound source localization (SSL) and face detection and then applying a suitable attention shifting strategy to turn the robot towards the speaker(s). We have experimented with a TurtleBot2~\cite{turtlebot2}, which is equipped with a mobile base, 3D vision sensor and basic processing unit (Fig.~\ref{fig:turtlebot}). We have attached a raspberry-pi based microphone array with the bot for SSL, which is interfaced with a laptop. We have placed our robotic set up in a meeting like environment and allowed it to interact with the people present there. It has been observed that it is able to turn its attention naturally like a human to the speaking person almost all the time and retain focus on the speaking person until s/he has stopped talking and another person has started a conversation.

Our main contributions are twofold:
\begin{itemize}
	\item We have designed an audio-visual perception method (Section \ref{RP}) to accomplish this task. For sound source localization (SSL), we have used the time difference of arrivals (TDOA) technique. Further, it is coupled with face detection and lip movement detection to bring the speaker at the center of the viewing front in real-time.
	
	\item We also have defined rules for the robot (Section \ref{HRI}) to turn to a speaker in a group scenario where multi-speaker scenarios are also considered. We have proposed a state representation model (Section \ref{RSR}) for attention shifting between people in groups based on the defined rules.  The model is generic enough that it can be extended for other scenarios where interaction between human and robot is prevalent. 
\end{itemize}

We evaluate our system towards the participants' experience on the basis of likability and belongingness in the meeting scenario and satisfaction. Our study shows the usability of our system built atop TurtleBot2 from the user's perspective, experience, and evaluation.

\section{Survey of related works}
\label{Related}
In this section, we briefly discuss the existing works on sound source localization (SSL) for robotic applications. Furthermore, we review some of the works on multi-modal systems for speaker detection in robotic applications such as service robotics and robotic assistant. At the end of this section, we provide an overview of the telepresence robotics, which is also the target application of our work. 
\subsection{SSL in robotics:} SSL finds application in many domains including smart conference system, robotics, and navigation. Rascon \textit{et al.}~\cite{rascon2017ssl} puts forward an end-to-end robotic system (implemented atop Golem-II+) that makes use of sound source localization to detect the direction of the speaking person. A hotel scenario has been considered in the work and the robot is potentially replacing a waiter on duty. Multi-DOA estimation method has been used for SSL and three strategies have been incorporated -- locating distant speaker using SSL techniques, turning towards the customer while taking orders, and announcing out its inability to detect speaker direction due to multiple speakers speaking at a time or ambient noises. 
Similarly, Bennewitz \textit{et al.}~\cite{bennewitz2005guide} proposes a multi-modal HRI system for a humanoid robot that makes use of visual perception, SSL, and speech recognition together to shift attention between speakers in group talks. It is being used in a museum. The robot can use hand and eye gestures to direct the attention of the communicating person towards displayed items. However, such a audio-visual perception system for attention shifting needs a formal state representation modal for proper execution. The article lacks such a model description. 

There are some studies that tried to detect the speaker(s) based on visual perception. Bendris \textit{et al.}~\cite{bendris2010lip} have proposed a method of separating speaking faces from non-speaking faces in videos using lip activity detection. They have used a method to determine disorder in pixel direction and depending on the amount of disorder, they have detected the speaking faces. They have achieved high classification accuracy in their results. Similarly, Joosten \textit{et al.}~\cite{joosten2015facial} have proposed a method for detecting voice activity in video using facial movement calculations. Spatio-temporal Gabor filters (STem-VVAD) have been used in their method to separate speech from non-speech frames. They have used the models in the face region of the frame and achieved beneficial results.   

Vazquez \textit{et al.}~\cite{vazquez2016focus} discussed a reinforcement learning based method to find strategies for shifting attention of the robot to the speaking person in a group conversation. The article also showed an interesting state representational model for formulating good policies for attention shifting in a robot in such a scenario. This work was extended in their future article~\cite{vazquez2017autonomy}, where they used perception of robot regarding a speaker in a group of 2 to 3 people and how accurately it could shift attention to the speaker in focus. The article also discussed social aspects in HRI where the gaze behaviour of a robot can determine a person's perception of the robot's motion and vice-versa. So, robot's motion and gaze must be co-designed rather than individually dealing them. Additionally, the article shed lights on the people's view how well they feel about the robot's belonging to the group considering the gaze, orientation, and motion of the robot.  

In this article, we have attempted to propose a formal state representation model in general, which is more suitable for a meeting scenario (shown in Section \ref{RSR}). Our model is capable of handling any interaction (specifically attention shifting) scenario with slight modification and customization.

\subsection{Telepresence robotics:}
\label{TPR}
Since our main target application is a telepresence robot, we briefly discuss about the related works in telepresence robotics domain. Venolia \textit{et al.}~\cite{venolia2010embodied} presented their telepresence device (Embodied Social Proxy) to enable a satellite co-worker to be present in the meetings with the collocated co-workers. This improves the interpersonal connection socially. They suggest that since their device is not mobile, only limited features are available and this remains a future research direction. Perhaps a mobile robotic system with the existing features can be beneficial. We have tackled this problem by designing our HRI scheme atop TurtleBot2 (shown in \ref{fig:turtlebot}). Biehl \textit{et al.}~\cite{biehl2015not} discussed the adverse effects of using Embodied Social Proxy devices similar to \cite{venolia2010embodied}. They concluded that the remote participant attending a meeting using such a device tends to create confusion among the speakers, e.g., when to speak. This happens due to the lack of sense of presence and belongingness for the remote participant. Some recent works \cite{neustaedter2016beam, rae2017robotic} studied the use-case of using telepresence robots to attend conferences remotely. They placed many recommendations for future research in such systems, one of them being an audio-visual feed of the conference (live) to the remote attendee/participant with ease and better viewing angles.

Stoll \textit{et al.}~\cite{stoll2018wait} presented a study towards the shortcomings of the current state of the art in telepresence robotics focusing on the participation ease of remote and collocated attendees' perception of the robot and the remote attendee. They experimented with a collaborative task to solve a word puzzle where two collocated attendees team up with a remote attendee using a telepresence robot. It reveals that there is sufficient in-balance of collaborative behaviour between collocated attendees and remote attendee. The collocated attendees tends to interact with other collocated attendees more than they do with a remote attendee. The main reasons for this being the lack of ease of interaction among the collocated attendees and remote attendee through the robot. Recent works lack in viewing accuracy due to limited viewing area or face-to-face interaction, lack of eye gazes, gestures, body movement, attention shifting strategy, and body language. In a nutshell, there is a lack of feeling of belongingness in the remote attendee or participant about his/her presence in the actual location and same feeling goes for the collocated attendees/participants in the actual location about the remote participant. 

In this article, we address the recommendation of various studies using an audio-visual perception system (Sections \ref{RP}) and a state representation model to control that system (Section~\ref{RSR}). We partially attend to the problem of face-to-face interaction and attention shifting using an audio-visual model. Moreover, we have made an attempt to improve the viewing angle and area by shifting attention on the visual input as and when required (guided by HRI rules in Section~\ref{HRI}). This improves the sense of presence and belongingness for the remote attendee/participant.

\section{Perception system: Overview and design}
\label{Design}

A telepresence robot is used to attend meetings on someone's behalf. The robot can be used to stream the audio and video of the meeting proceedings through the microphone and camera integrated into the robot. Though standard video conferencing systems are widely used to attend such meetings remotely, the camera in such a system is adjusted to a certain viewing angle (maximum 150$^{\circ}$-180$^{\circ}$). This provides a video feed of a fixed area to the remote participant. This also results in limited capturing of the activities happening in the meeting. Moreover, the camera has to be adjusted accordingly. Although there are some video-conferencing systems equipped with cameras which can rotate or adjust themselves as per need but robust rotation strategies are not in place. Such cameras also lack flexibility issues as they are fixed into some table or platform. Therefore, to provide the entire proceedings of the meeting to the remote participant, either the activities have to be restricted within the limited viewing area or the camera has to be adjusted. On the other hand, if a robot is capable enough to understand the meeting environment and detect the speakers and activities of the speakers using audio-video perception, then it will be an added advantage and least human intervention will be required. This can also provide a 360$^{\circ}$ view of the meeting proceedings, where the robot captures video of the speaker and the activities as and when that person is the center of attention. The audio/video feed can be stored as well for further reference and analysis. This kind of a robotic system can also be further extended for more complicated telepresence scenario where the proceedings are not limited to a particular room. The basic requirement here is to identify different speakers in a meeting scenario and adjust the robot's pose to bring focus on the current speaker. Although \cite{rosebrock} presents a video conferencing environment with eye gaze perceiving robot, but it can only determine whom to look or who is going to speak on the basis of other attendees' eye gaze direction. But such a system alone cannot perfectly replicate a remote person's behaviour until a multi-modal audio-visual system is used for gaining perception.

Our multi-modal perception system is a combined system using SSL and visual (face detection and lip movement detection) system. An audio-visual hybrid system is being incorporated to achieve the best possible interaction experiences. We have used a Quantum QHM495LM 25MP web camera for visual perception design and a ReSpeaker 4-Mic Array \cite{respeaker4mic} for sound localization and DOA estimation using Voice Activity Detection (VAD) and Time Difference of Arrival (TDOA) \cite{rascon2017ssl} methods. The ReSpeaker 4-Mic Array is mounted atop Raspberry-pi 2 computer board \cite{rpi} for collecting the audio data and processing it using TDOA method. We choose TDOA method for SSL because it is one of the least computational costly and provide fast DOA estimation in real-time with acceptable accuracy. Though there are more accurate algorithms such as MUSIC \cite{rascon2017ssl,schmidt1986multiple}, they are computationally costly and using them in real-time (compromising their filters and frequency band) degrade the accuracy to a considerable extent. So, using such algorithms in real-time application such as robot sound source localization, without degrading the accuracy much, remains a scope of future research.  

\begin{figure}
	\includegraphics[width=1\linewidth]{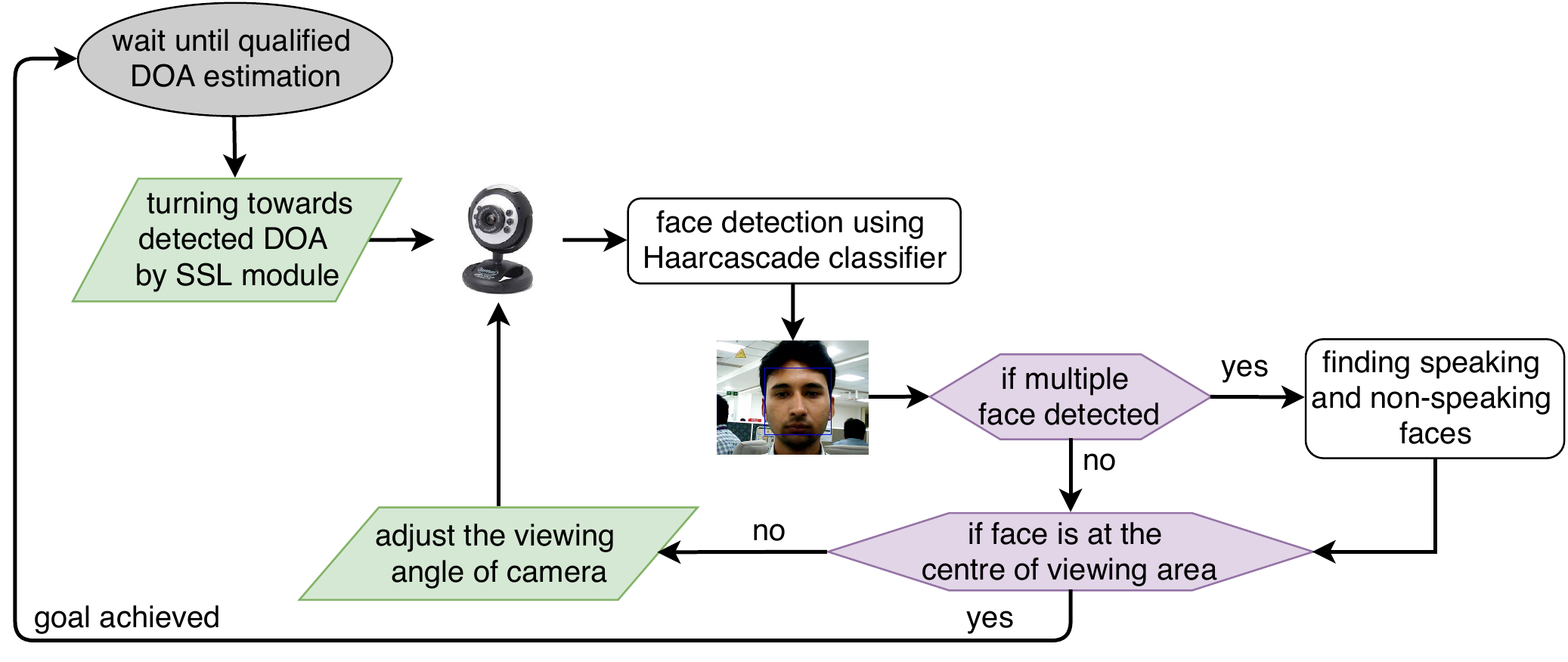}
	\centering
	\caption{Overview of the visual perception and rotation control system.}
	\label{fig:camera_control}
	\vspace{-6.1mm}
\end{figure}

\begin{figure}
	\includegraphics[width=0.8\linewidth]{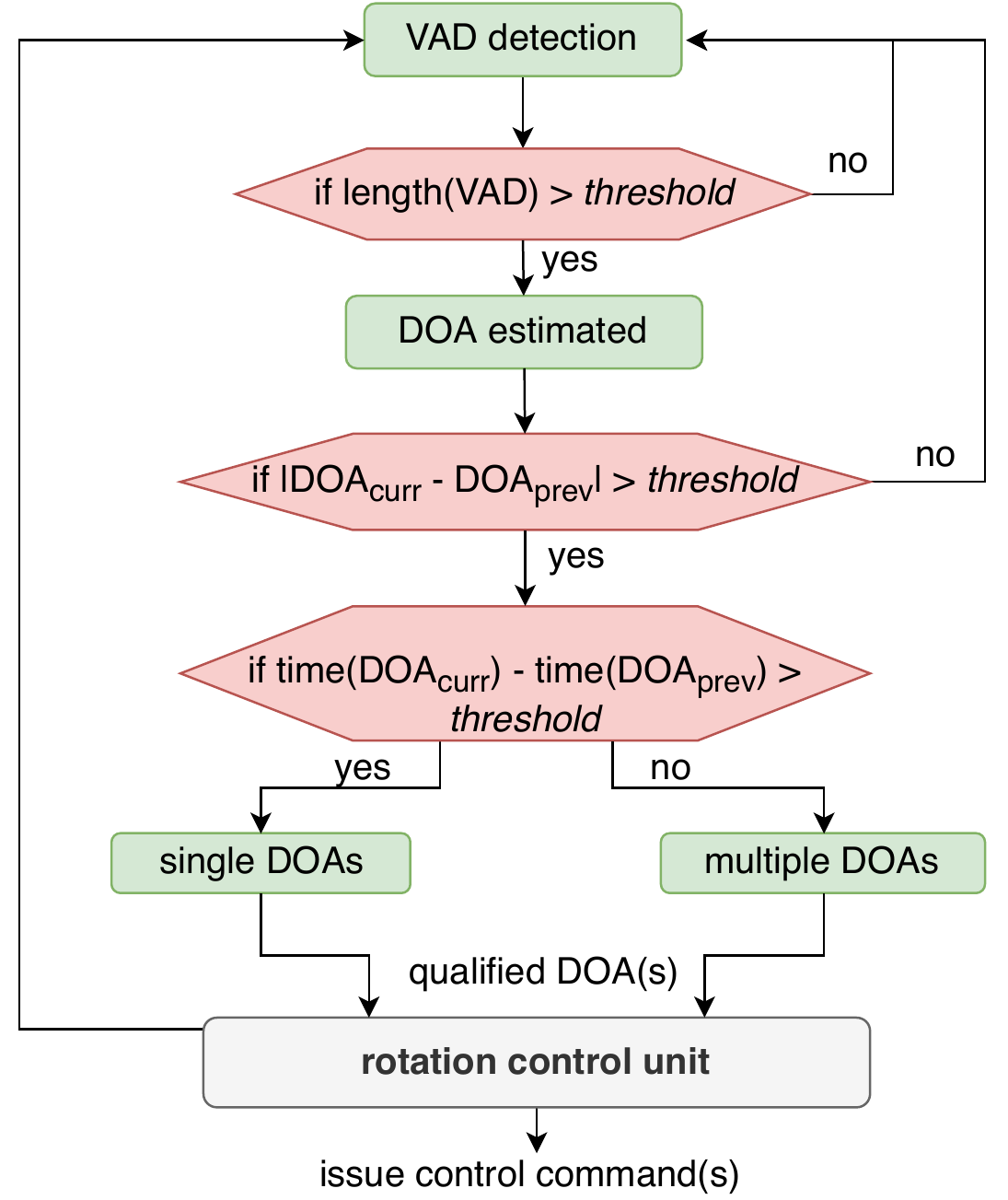}
	\centering
	\caption{Flowchart: Rules to find qualified DOAs in SSL.}
	\label{fig:three}
	\vspace{-7.2mm}
\end{figure}

\subsection{Robot perception}
\label{RP}

Our robot is capable of attending a meeting with a number of attendees. It has a web camera and a microphone array of 4 mics. The mic array is used to calculate DOA using TDOA method. After estimation of the DOA for any speaker in the meeting, the robot may turn to the speaker on the basis of the HRI rules (for extracting qualified DOAs) formulated in Section~\ref{HRI} and State representation model discussed in Section~\ref{RSR}. The second part is the visual perception where the robot adjusts its gaze to a particular speaker in view using face detection. If there are multiple faces in its view, it further detects lip movement and determines the speaker in the current frame. The above mentioned processes go on executing as and when qualified DOAs (Section~\ref{HRI}) are being estimated. Fig.~\ref{fig:turtlebot} shows a typical setup for estimating DOA using ReSpeaker 4-Mic Array and Raspberry-pi 2 computer board.

The visual system is being shown in Fig.~\ref{fig:camera_control}. The video input is streamed using the Quantum QHM495LM 25MP web camera, which is then fed into a Haar cascade classifier for face detection~\cite{tiwari}. If a face is detected in the viewing area of the camera, the robot will automatically try to rotate and adjust itself till the face is exactly in the middle of the view. We have achieved this using boundary pixels of the surrounding box of the detected face. The robot is allowed to adjust until the boundary pixels of the surrounding box is exactly in the middle of the view. If the number of faces detected in one view is more than 1, we apply lip movement detection methods \cite{rosebrock2017detect,agarwal2018smilefie} to identify the speaking person among the detected faces. Then the robot rotates and self adjusts to bring the speaking face to the center of the view. If at any point of time, more than 1 faces are detected with lips movement, the robot will place its attention to a middle point (neutral area) of all the detected faces with lips movement. The lips movement detection atop the face detection technique is shown in Fig.~\ref{fig:lips_move}. Figures \ref{fig:four_1}, \ref{fig:four_2}, and \ref{fig:four_3} show the shift of focus when a speaking face is detected in a group of detected faces. Fig.~\ref{fig:four_4} shows the focus in the middle of the group if all the detected faces are speaking.

\subsection{Robot state representation}
\label{RSR}
Apart from the audio-visual perception system implemented in the paper, we also propose a state representation modal for the robot in a group meeting scenario. We adapt the state representation model proposed in~\cite{vazquez2016focus} for our robot. This representation is well suited for the meeting scenario and can be extended for other interaction scenarios in the future.

Any state of the robot during interaction in the meeting room can be represented with some parameters as shown below. We define 8 parameters, having 4 action parameters and 4 control parameters. Parameters; \textit{$\theta_{1}$}, \textit{$\theta_{2}$}, \textit{$\theta_{3}$} and \textit{$\theta_{4}$} are action parameters which values determine the next pose of the robot (in terms of rotation angle). And, parameters; \textit{f\textsubscript{1}}, \textit{f\textsubscript{2}}, \textit{f\textsubscript{3}} and \textit{f\textsubscript{4}} are the control parameters which values determine which actions or action parameters are to be considered at any point of interaction.       
Therefore, a state of a robot at any point of the interaction may be represented as a tuple, as shown below:
\begin{align*}
\textit{S}(\theta_1, f_1, \theta_2, f_2, \theta_3, f_3, \theta_4, f_4)
\end{align*}

\textit{$\theta_{1}$} denotes the parameter having values in [0, 360], which represents the rotation that the robot needs to take in order to direct its attention towards the speaking subject (as per estimated DOA) using SSL.

\textit{f\textsubscript{1}} denotes the parameter having values in [0, 1], which denotes if in any point of time \textit{$\theta_{1}$} is valid or not. It depends upon the detection of Voice Activity (VAD) . If voice activity is detected, it is set to 1, (\textit{$\theta_{1}$} is valid) else 0 (\textit{$\theta_{1}$} is invalid).

\textit{$\theta_{2}$} denotes the parameter valuing between [0, 360], which denotes the rotation, that needs to be done to adjust the speaker to put it in the center of the robot's viewing angle using visual perception and face detection.

\textit{f\textsubscript{2}} denotes the parameter having values in [0, 1] determining the validity of parameter \textit{$\theta_{2}$}. It assumes value 0 if the speaker is already in the center of the viewing area (means \textit{$\theta_{2}$} is invalid) else it is 1. If \textit{f\textsubscript{2}} is 1, it resets \textit{f\textsubscript{1}} to 0 until next qualified DOA is estimated (for turning) as per our defined rules (HRI rules discussed in Section \ref{HRI}). 

\textit{$\theta_{3}$} denotes the parameter having values in [0, 360], which denotes the rotation which the robot takes in case multiple DOAs are detected in a short time interval or speech activity is detected. In such case, it should turn to a neutral location or to the average DOA of the previously detected DOAs in that region or cluster (speech activity and region/cluster are defined in Section \ref{HRI}).

\textit{f\textsubscript{3}} denotes the parameter having values in [0, 1] determining the validity of parameter \textit{$\theta_{3}$}. It assumes the value 0 if multiple DOAs are not detected in a given small time frame or speech activity (speech activity is defined in Section \ref{HRI}) is not detected (i.e., parameter \textit{$\theta_{3}$} is invalid) else the value is 1. If \textit{f\textsubscript{3}} is 1, it automatically resets \textit{f\textsubscript{1}}  and \textit{f\textsubscript{2}} to 0 until next qualified DOA is estimated for action as per the defined rules (HRI rules discussed in Section \ref{HRI}).   

\textit{$\theta_{4}$} denotes a parameter assuming values in [0, 360]. This parameter defines rotation or attention shifting between people detected in a single viewing area as and when one of them starts speaking. Lips movement detection is being used atop the detected faces to determine the speaking person and bring him/her to the center of the viewing area. If more than one faces are detected with lips movement in any point of time, the focus is placed in the middle area of all the detected faces with lips movement. 

\textit{f\textsubscript{4}} denotes a parameter assuming values in [0, 1]. This parameter defines the validity of \textit{$\theta_{4}$}. It assumes value 0 (i.e. invalidates parameter \textit{$\theta_{4}$}) when only one face is detected in a robots viewing area using \textit{$\theta_{1}$} and \textit{$\theta_{2}$}. In case of \textit{$\theta_{3}$}, it remains 0 until someone from the group starts speaking. In rest of the cases it is 1. When it assumes 1, it automatically resets \textit{f\textsubscript{1}}, \textit{f\textsubscript{2}} and \textit{f\textsubscript{3}} to 0 until next qualified DOA is estimated for action as per the defined rules (HRI rules discussed in Section \ref{HRI}).    

Hence, the robot is at \textquotesingle \textit{S\textsubscript{init}}\textquotesingle\phantom{ }in the beginning of the interaction where all the control parameters are 0. Hence, all the action parameters must be essentially invalid. After a VAD is detected, leading to DOA estimation for speaker, it goes to a new state \textquotesingle \textit{S\textsubscript{new}}\textquotesingle \phantom{ }as per the above defined state representation model. Subsequently state changes to new states as and when new qualified DOAs are estimated as per the defined rules (HRI rules discussed in Section \ref{HRI}). For the robot to go from one state to another at least one of the control parameters must be 1 i.e. one of the action parameters must be essentially valid at any point in time. \textit{Algorithm 1} shows the procedure for state changes in our proposed robot state representation model.

\begin{algorithm}
	\begin{footnotesize}
	\begin{algorithmic}[0.7]
		\STATE  \textbf{State} = \textit{S\textsubscript{init}}\par
		\STATE \textit{f\textsubscript{1}}, \textit{f\textsubscript{2}}, \textit{f\textsubscript{3}}, \textit{f\textsubscript{4}} = 0;\par
		\STATE \textit{$\theta_{1}$}, \textit{$\theta_{2}$}, \textit{$\theta_{3}$}, \textit{$\theta_{4}$} are invalid;\par
		\lIf{VAD detected};
		\STATE Estimate DOA as per rule;\par
		\STATE \textit{f\textsubscript{1}} = 1;\par
		\textit{$\theta_{1}$} is validated;\par
		\STATE \textbf{State} = \textit{S\textsubscript{new}}\par
		\lIf{\textit{f\textsubscript{3} is still 0}};
		\hspace{1ex} \lIf{\textit{f\textsubscript{2} is still 0}};
		\hspace{1ex}\hspace{1ex}  \lIf{\textit{f\textsubscript{4} is still 0}};
		\hspace{1ex}\hspace{1ex}\hspace{1ex}  \STATE Target \textbf{State} reached until next qualified DOA estimated;\par
		
		\hspace{1ex}\hspace{1ex}  \Else{\STATE\hspace{1ex}  \textit{$\theta_{4}$} is validated;\par \hspace{1ex} \Repeat{f\textsubscript{4} is 0}{\STATE \textbf{State, \textit{S\textsubscript{new}}} = \textit{S\textsubscript{new+1}};\par
				\STATE  \textit{f\textsubscript{1}},\textit{f\textsubscript{2}},\textit{f\textsubscript{3}} = 0;\par  } \STATE Target \textbf{State} reached until next qualified DOA estimated;\par
		}
		\hspace{1ex} \Else{\STATE \textit{$\theta_{2}$} is validated;\par
			\Repeat{f\textsubscript{2} is 0}{
				\STATE \textbf{State, \textit{S\textsubscript{new}}} = \textit{S\textsubscript{new+1}};\par
				\STATE  \textit{f\textsubscript{1}} = 0;\par  
				
			}
			\hspace{1ex}  \hspace{1ex} \lIf{\textit{f\textsubscript{4} is still 0}};
			\hspace{1ex}  \hspace{1ex}\hspace{1ex} \STATE Target \textbf{State} reached until next qualified DOA estimated;\par
			\hspace{1ex} \hspace{1ex} \Else{\STATE \hspace{1ex} \textit{$\theta_{4}$} is validated;\par\hspace{1ex} \Repeat{f\textsubscript{4} is 0}{\STATE \textbf{State, \textit{S\textsubscript{new}}} = \textit{S\textsubscript{new+1}};\par
					\STATE  \textit{f\textsubscript{1}},\textit{f\textsubscript{2}},\textit{f\textsubscript{3}} = 0;\par  }\STATE Target \textbf{State} reached until next qualified DOA estimated;\par
			}
		}

		\Else{\STATE \textit{$\theta_{3}$} is validated;\par
			\Repeat{f\textsubscript{3} is 0}{
				\STATE \textbf{State, \textit{S\textsubscript{new}}} = \textit{S\textsubscript{new+1}};\par
				\STATE  \textit{f\textsubscript{1}}, \textit{f\textsubscript{2}} = 0;\par  
				
			}
			Qualified DOA estimated from one person in group;\par
			\hspace{1ex} \hspace{1ex} \lIf{\textit{f\textsubscript{2} is still 0}};
			\hspace{1ex} \hspace{1ex} \hspace{1ex} \lIf{\textit{f\textsubscript{4} is still 0}};
			\hspace{1ex}  \hspace{1ex}\hspace{1ex}  \hspace{1ex}\STATE Target \textbf{State} reached until next qualified DOA estimated;\par
			\hspace{1ex}\hspace{1ex} \hspace{1ex} \Else{\STATE \hspace{1ex} \textit{$\theta_{4}$} is validated;\par\hspace{1ex} \Repeat{f\textsubscript{4} is 0}{\STATE \textbf{State, \textit{S\textsubscript{new}}} = \textit{S\textsubscript{new+1}};\par
					\STATE  \textit{f\textsubscript{1}},\textit{f\textsubscript{2}},\textit{f\textsubscript{3}} = 0;\par  } \STATE Target \textbf{State} reached until next qualified DOA estimated;\par \Else{\STATE \textit{$\theta_{2}$} is validated;\par
					\Repeat{f\textsubscript{2} is 0}{
						\textbf{State, \textit{S\textsubscript{new}}} = \textit{S\textsubscript{new+1}};\par
						\STATE  \textit{f\textsubscript{1}} = 0;\par  
						
					}
					\hspace{1ex}  \hspace{1ex} \lIf{\textit{f\textsubscript{4} is still 0}};
					\hspace{1ex}  \hspace{1ex}\hspace{1ex}  \STATE Target \textbf{State} reached until next qualified DOA estimated;\par\par
					\hspace{1ex} \hspace{1ex} \Else{\STATE \hspace{1ex} \textit{$\theta_{4}$} is validated;\par\hspace{1ex} \Repeat{f\textsubscript{4} is 0}{\STATE \textbf{State, \textit{S\textsubscript{new}}} = \textit{S\textsubscript{new+1}};\par
							\STATE  \textit{f\textsubscript{1}},\textit{f\textsubscript{2}},\textit{f\textsubscript{3}} = 0;\par  }\STATE Target \textbf{State} reached until next qualified DOA estimated;\par
					}
				}
			}
		}
		\caption{Procedure for state change in our Robot for attention shifting}
		\label{algo:b}
	\end{algorithmic}
	\end{footnotesize}
\end{algorithm}

\section{Human-Robot Interaction rules for our robot: Bringing naturality in attention shifting activity}
\label{HRI}
The HRI module in our system consists of an audio-visual multi-modal perception design. Firstly, the SSL module does the sound source localization in real-time and then the vision module adjusts the robot's attention to the exact speaker as described in Section \ref{RP}. 

We have incorporated rules in the SSL module for efficient attention shifting of the robot towards the desired speaker in a meeting scenario.

Two definitions to be considered for the rules:

\textbf{Speech activity:} If the length of VAD (continuous sound signal received without break) is greater than a defined threshold, we call it as Speech activity.

\textbf{Region or Cluster:} Group of attendees seating nearby each other in a meeting scenario. We have calculated this using distance between estimated DOAs. We have formed clusters of attendees in the meeting scenario considering that initially a cluster contains 1 DOA. As and when new DOA is estimated, it is dynamically assigned to an existing region or cluster depending on its distance from the center (average DOA) of the existing clusters, then the new average is recalculated. This method is adapted from the work of Rascon \textit{et al.}~\cite{rascon2015hri}.

\begin{figure*}
	\centering
	\subfigure[one speaker (Rule 1 or 3)]{\includegraphics[width=0.2\linewidth]{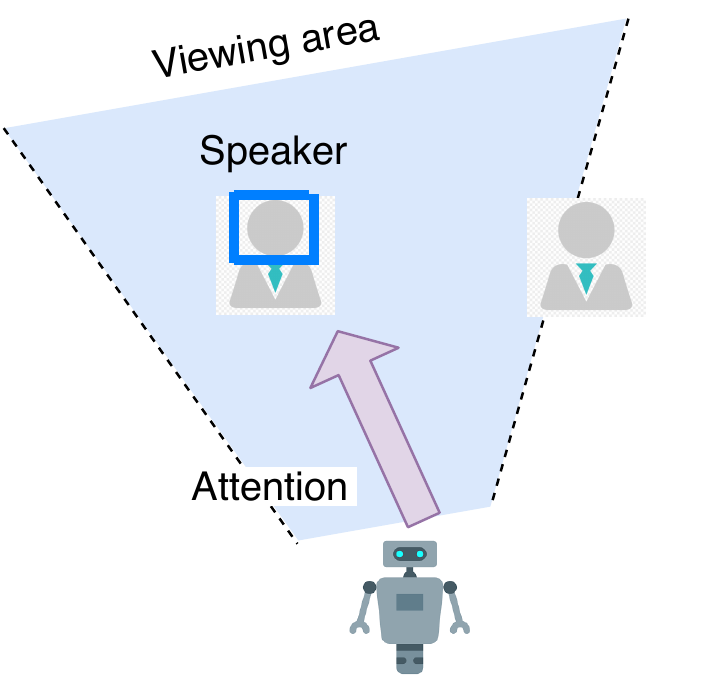}\label{fig:one_speaker}}
	\hspace{0.8cm}
	\subfigure[two speakers (Rule 2)]{\includegraphics[width=0.2\linewidth]{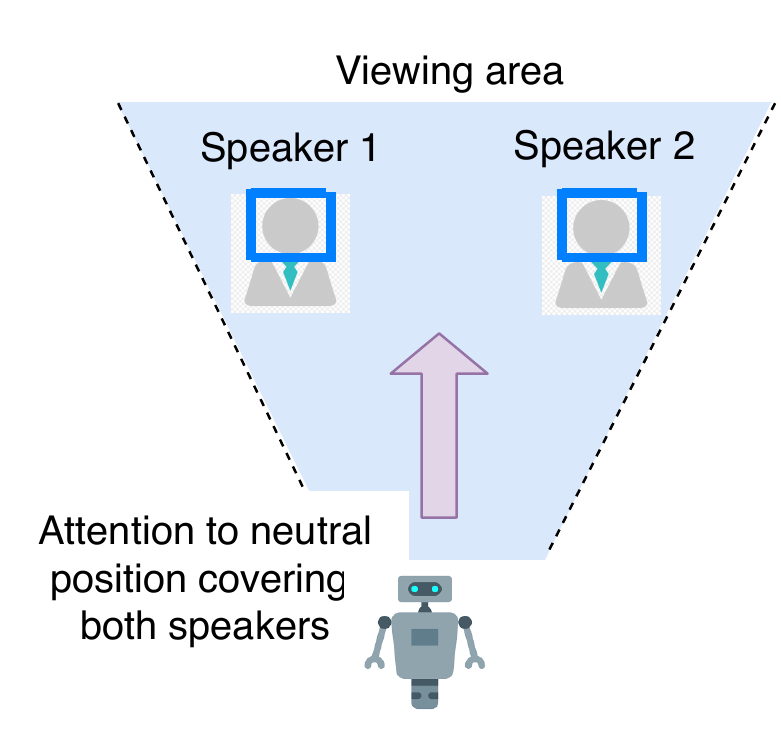}\label{fig:two_speakers}}
	\hspace{0.8cm}
	\subfigure[multiple speakers (Rule 4)]{\includegraphics[width=0.2\linewidth]{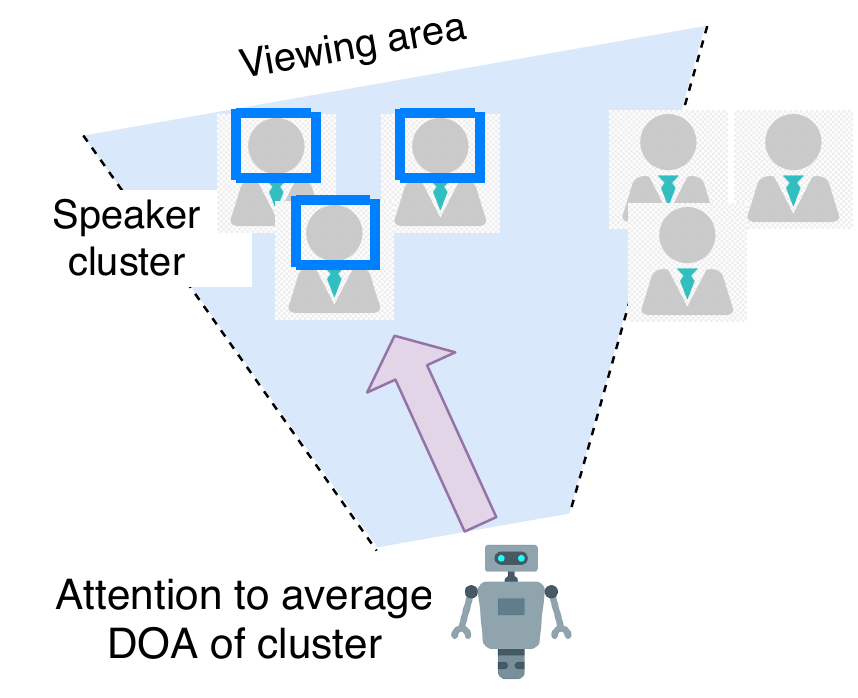}\label{fig:multi_speakers}}
	\caption{In a meeting with multiple person, how the robot shifts attention based on speakers' location.}
	\label{fig:attention}
\end{figure*}

\begin{itemize} 
	\item \textit{\textbf{Rule 1}}: Turn only if there is a gap of more than 2 seconds between two consecutive DOA estimation (Fig.~\ref{fig:one_speaker}).  
	\item \textit{\textbf{Rule 2}}: Turn to a neutral location if the time gap between two consecutive DOA estimations is less than 2 seconds i.e multiple DOAs estimated within a short time frame (Fig.~\ref{fig:two_speakers}).
	\item \textit{\textbf{Rule 3}}: Turn only if the angle difference between two consecutive DOA estimations is greater than 5 (Fig.~\ref{fig:one_speaker}).
	\item \textit{\textbf{Rule 4}}: Turn to the average DOA of a group of attendee (or region/cluster) if speech is detected from that group (Fig.~\ref{fig:multi_speakers}).
\end{itemize}

The Fig.~\ref{fig:three} is a more generalized representation of the flow of SSL program with the incorporated rules. Here, the thresholds are estimated after extensive experimentation, to best suit the expectations of the users and bring naturality in interactions. The output of such a rule-based SSL system is the qualified DOAs which will be further considered by the visual system (discussed in Section \ref{RP}) and state representation model (discussed in Section \ref{RSR}) for determining the rotation angle for the robot and issue the control command as per the current situation.

As already stated above, Figures \ref{fig:one_speaker}, \ref{fig:two_speakers} and \ref{fig:multi_speakers} show the different scenario of attention shifting as per the above defined rules.

\begin{figure}
	\centering
	\subfigure[Lips movement detection in two faces detection scenario.]{\includegraphics[width=0.22\linewidth]{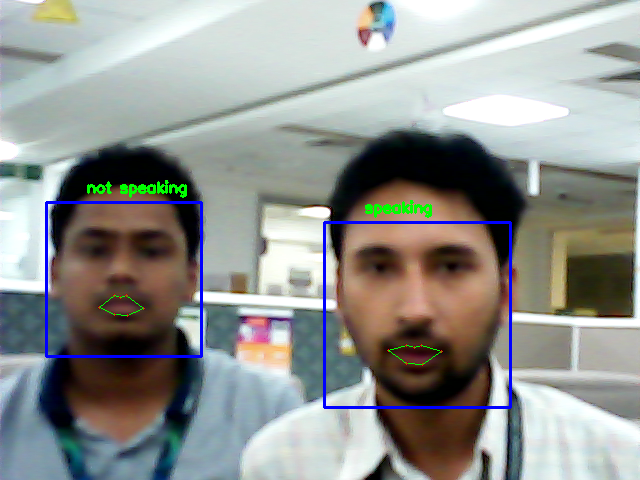}\label{fig:four_1}}
	\hspace{0.05cm}
	\subfigure[Shifting gaze to the speaking person ; self adjust]{\includegraphics[width=0.22\linewidth]{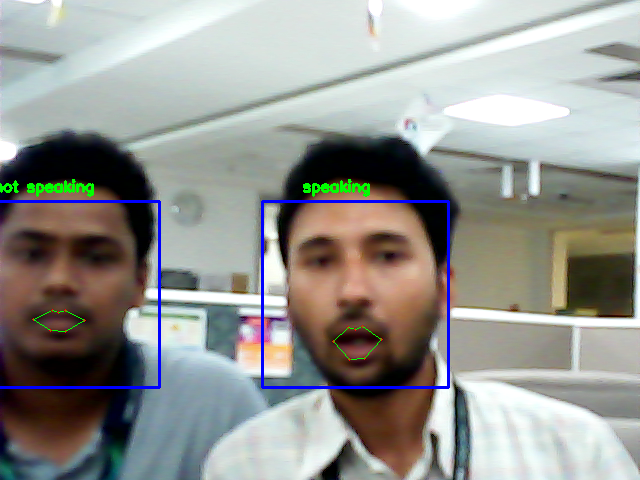}\label{fig:four_2}}
	\hspace{0.05cm}
	\subfigure[Positioning the speaking person in the middle of the viewing window.]{\includegraphics[width=0.22\linewidth]{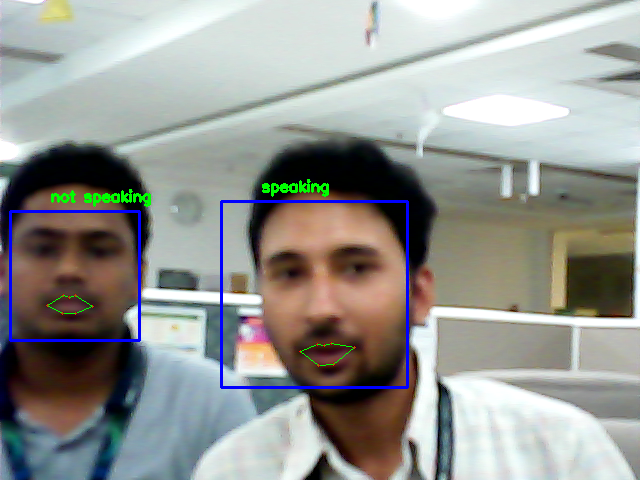}\label{fig:four_3}}
	\hspace{0.05cm}
	\subfigure[Multiple speaking faces detected using lips movement detection.]{\includegraphics[width=0.22\linewidth]{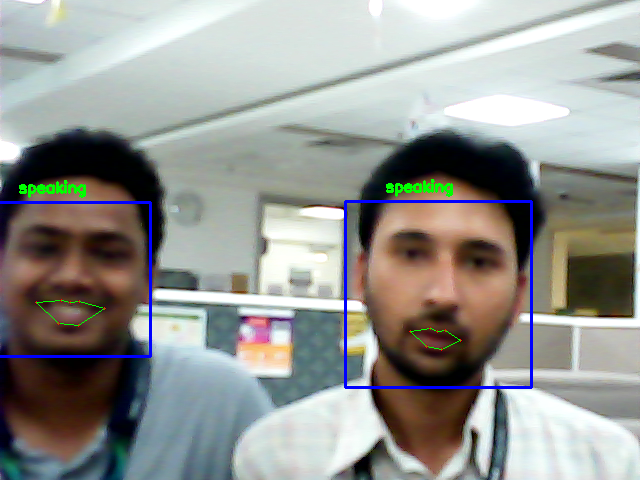}\label{fig:four_4}}
	
	\caption{In a multiple face detected scenario, lips movement detection plays an important role in identifying a speaking person (generally when Rule 2 or 4 followed in SSL module).}
	\label{fig:lips_move}
\end{figure}

The Figures \ref{fig:four_1}, \ref{fig:four_2},  \ref{fig:four_3} and \ref{fig:four_4} are the depiction of the visual perception (including lips movement detection) discussed in Section \ref{RP}. This takes place generally after Rule 2 or 4 is followed from the above defined rules. In case of Rule 1 generally only one face is detected (so lips movement detection is not required), and the robot adjusts its focus to bring it to the center of the viewing area (also discussed in Section \ref{RP}). 

\section{Behavioural evaluation: Presenting our robotic avatar to the meeting attendees}
\label{ERD}


\begin{figure}[ht]
	\includegraphics[width=\linewidth]{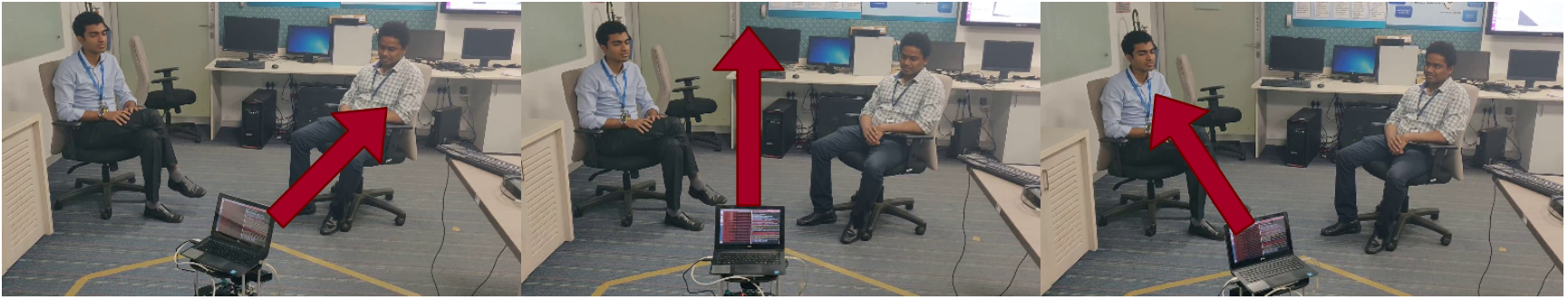}
	\centering
	\caption{Our robotic system with two collocated attendees in meeting scenario.}
	\label{fig:ten}
\end{figure}

Our goal is to provide a better HRI scheme (specifically attention shifting model) for the remote attendee to communicate with the collocated attendees and vice-versa in a meeting scenario. To achieve this goal we propose the perception system (Section \ref{RP}) and the robot state representation (Section \ref{RSR}). Moreover, we define some rules to get qualified DOAs (Section \ref{HRI}). We also have partially attended to drawbacks of the current state of the art research and some of the future research possibilities as mentioned in  \cite{stoll2018wait}. We also adapt some of the existing concepts from \cite{vazquez2016focus} (for state representation) and \cite{rascon2015hri} (for cluster formation in multi- DOA scenario).

\subsection{Evaluation methodology}
\label{EM}
We have defined some parameters to evaluate the system based on the user's experience and feedback. Table \ref{tab:zero} has described 5 parameters for the purpose. We have asked each of the participants (collocated attendees and remote attendees) to first go through these parameters. After that, we have asked them to rate their experience (on a scale of 1-10), regarding ease of the interaction and feeling of belongingness, for these five parameters. We call the final calculated value as User Experience/Satisfaction Index (\textit{UEI}). The higher the value of \textit{UEI}, the higher is the likability and usability of our system. The higher the values of the defined parameters, the higher is the \textit{UEI} value. We asked the participants to rate each parameter on a 1-10 scale based on the below given logic for each interaction. 
\begin{itemize} 
	\item \textit{{p\textsubscript{1}}} will be higher if lower unnecessary turns are included in that interaction session.
	\item \textit{{p\textsubscript{2}}} will be higher if lower necessary turns are missed in that interaction session.
	\item \textit{{p\textsubscript{3}}} will be higher if the robot turns to the actual speaker with lower unacceptable accuracy in that interaction session.
	\item \textit{{p\textsubscript{4}}} will be higher if the robot has misjudged the speaking person lower number of times in that interaction session.
	\item \textit{{p\textsubscript{5}}} will be higher if the robot has missed the entire process of speaker detection lower number of times in that interaction session.
\end{itemize}
Then, we calculate \textit{UEI}, giving equal weightage to each of the parameters, as follows:

\begin{equation} \label{eq:someequation}
UEI={\dfrac{\textit{p\textsubscript{1}}+\textit{p\textsubscript{2}}+\textit{p\textsubscript{3}}+\textit{p\textsubscript{4}}+\textit{p\textsubscript{5}}}{5}}   
\end{equation}
Our evaluation model can be extended to other interaction, specifically attention shifting systems for evaluating its likability, user experience, usability, and satisfaction.
New parameters can be added to make the evaluation process more compelling and suited for other scenarios.
We have allowed the robot to interact in two scenarios:

\subsection{Robot and two collocated attendees}
We placed the robot in a meeting setup with two collocated attendees in front of it as shown in Fig.~\ref{fig:ten}. The remote attendee controlled the robot from a remote machine. We allowed an interaction of 3-6 minutes. The collocated attendees interacted within themselves and the remote attendee through the robot. We took feedback from the collocated attendees. They accepted the usefulness of the system and were impressed by its perception of speaking person or attention shifting capabilities. They gave an average \textit{UEI} of \textbf{7} in the scale of 1-10 (\textit{UEI} calculated using the equation 1). We also took feedback from the remote attendee, who also expressed his/her satisfaction by giving a \textit{UEI} of \textbf{7}. Then we asked about the collocated attendees' feelings on the robot's participation in the interaction (how well they feel that the remote attendee is present there itself) and remote attendee's feeling on his/her belongingness in the meeting. They responded positively and recommended our system for any meeting scenario confined in a room. The audio-visual perception system controlled by the robot state representational model has been able to perceive the environment and shift focus to the speaking person and center of attraction in an efficient and fast manner.           

\subsection{Robot and multiple collocated attendees (more than 2)} We then placed our robot in a scenario having more than 2 attendees as shown in Fig.~\ref{fig:multi_speakers}. We got similar positive responses from the collocated attendees and remote attendees. They rated our system with a \textit{UEI} of \textbf{7} (rounded off) out of 1-10 scale. In the case of multiple collocated attendees, the amount of belongingness was not impacted due to the visual system we have used. The robot can self adjust without remote attendee's intervention. The focus can be fixed to a single talking person or to the center of the group or cluster as per the rules we have defined and visual perception module. So, the remote attendee feels that he himself/herself is present in the meeting personally. Similarly, the collocated attendees, feel that the remote person or attendee is present there in the meeting as the robot behaves some what in a way a person behaves when he is in a meeting like an avatar of the real person. For example, a person shifts focus if some one speaks, then fix focus on a person in a group if s/he starts speaking, looks at a neutral position if two persons talk to each other or to the robot and focuses both person together and so on. Our robot is designed to address all such cases that can arise in a meeting. So, we are able to address some of the recommendations for future research and design suggested by Stoll \textit{et al.}~\cite{stoll2018wait} regarding increasing visual perception and feeling of belongingness of remote attendee in the meeting from collocated attendees' perspective and similar feeling of belongingness of the remote attendee as well. We have used TurtleBot2 as shown in Fig.~\ref{fig:turtlebot} for all of our experiments in meeting setups. The main limitation of our system atop TurtleBot2 is its height. So, to focus and detect faces, we have used an adjustable camera in the robotic avatar by putting its focus in an upper angle to view people's face seating in a meeting scenario. 

\begin{table}
	\fontsize{7.5}{7.5}\selectfont
	\centering
	\caption{Evaluation parameters : User satisfaction ratings between 1-10 scale on the basis of this table}
	\begin{tabular}{|p{3cm}|p{5cm}|} 
		\hline
		\bf{Parameters} & \bf{Description}\\
		\hline
		Unnecessary turns (\textit{\textbf{{p\textsubscript{1}}}}) & How many unnecessary turns are included in a single interaction session   \\
		\hline 
		Necessary turns (\textit{\textbf{{p\textsubscript{2}}}}) & How many necessary turns are missed in a single interaction \\
		\hline
		Turn accuracy (\textit{\textbf{{p\textsubscript{3}}}}) & How many times the   robot turns to the speaker with unacceptable accuracy \\
		\hline
		Attention accuracy in group of 2 or more people (\textit{\textbf{{p\textsubscript{4}}}}) & How many times the robot has misjudged the speaking person\\
		\hline
		Speaker detection accuracy (\textit{\textbf{{p\textsubscript{5}}}}) & How many times the robot has missed the entire process of speaker detection and attention shifting \\
		\hline
	\end{tabular}
	\label{tab:zero}
\end{table}

\section{Conclusions}
\label{CONC}
In this article, we present a study and implementation of an audio-visual multi-modal perception system for attention shifting to the target/speaking person in a meeting scenario where a robot attends the meeting on behalf of a remote attendee. The idea is to let a remote attendee attend the meeting with ease and feeling of belongingness in the meeting environment with least intervention (requiring to control the robot manually). Similarly, we consider the ease of interaction of the collocated attendees with the remote attendee through our (audio-visual) perception enabled robot (experimented with TurtleBot2). We use a real-time sound source localization and direction-of-arrival estimation module to give us the most appropriate directions of the speaker(s). We also make use of face detection and lip movement detection methods to adjust focus when there is a group of people. Additionally, we define a robot state representation model atop the audio-visual system to allow efficient state shifting (rotate to the required direction) at any point of time during the meeting. We have experimented with one, two and more collocated attendees and one remote attendee in each case. In all the scenarios, belongingness for both sides (remote and local attendees) is achieved through our robot model with high user satisfaction. 

In the future, we plan to extend this work by implementing multi-modal interaction on the robot's part using eye gaze perception and control and hand gesture systems so that the participants can understand the robot more precisely in meeting scenarios and vice versa. Moreover, reinforcement learning methods can be incorporated for defining more attractive and compelling strategies in attention shifting.

\balance
\bibliographystyle{acm}
\bibliography{main.bib}

\begin{thebibliography}{10}

\bibitem{rpi}
Raspberry pi.
\newblock \url{https://en.wikipedia.org/wiki/Raspberry\_Pi}.
\newblock From Wikipedia, the free encyclopedia.

\bibitem{respeaker4mic}
Respeaker 4-mic array for raspberry pi.
\newblock
  \url{http://wiki.seeedstudio.com/ReSpeaker\_4\_Mic\_Array\_for\_Ra\\spberry\_Pi/}.
\newblock Seeed Technology Co.,Ltd.

\bibitem{turtlebot2}
Turtlebot2.
\newblock \url{"https://www.turtlebot.com/turtlebot2/}.
\newblock Open Source Robotics Foundation, Inc.

\bibitem{agarwal2018smilefie}
{\sc Agarwal, R.}
\newblock Smilefie: how you can auto-capture selfies by detecting a smile.
\newblock
  \url{https://medium.freecodecamp.org/smilfie-auto-capture-selfies-by-detecting-a-smile-using-opencv-and-python-8c5cfb6ec197},
  7 August 2018.

\bibitem{bendris2010lip}
{\sc Bendris, M., Charlet, D., and Chollet, G.}
\newblock Lip activity detection for talking faces classification in
  tv-content.
\newblock In {\em International Conference on Machine Vision\/} (2010),
  pp.~187--190.

\bibitem{bennewitz2005guide}
{\sc Bennewitz, M., Faber, F., Joho, D., Schreiber, M., and Behnke, S.}
\newblock Towards a humanoid museum guide robot that interacts with multiple
  persons.
\newblock In {\em 5th IEEE-RAS International Conference on Humanoid Robots,
  2005.\/} (2005), IEEE, pp.~418--423.

\bibitem{biehl2015not}
{\sc Biehl, J.~T., Avrahami, D., and Dunnigan, A.}
\newblock Not really there: Understanding embodied communication affordances in
  team perception and participation.
\newblock In {\em Proceedings of the 18th ACM Conference on Computer Supported
  Cooperative Work \& Social Computing\/} (2015), ACM, pp.~1567--1575.

\bibitem{goodrich2008survey}
{\sc Goodrich, M.~A., Schultz, A.~C., et~al.}
\newblock Human--robot interaction: a survey.
\newblock {\em Foundations and Trends{\textregistered} in Human--Computer
  Interaction 1}, 3 (2008), 203--275.

\bibitem{joosten2015facial}
{\sc Joosten, B., Postma, E., and Krahmer, E.}
\newblock Voice activity detection based on facial movement.
\newblock {\em Journal on Multimodal User Interfaces 9}, 3 (2015), 183--193.

\bibitem{neustaedter2016beam}
{\sc Neustaedter, C., Venolia, G., Procyk, J., and Hawkins, D.}
\newblock To beam or not to beam: A study of remote telepresence attendance at
  an academic conference.
\newblock In {\em Proceedings of the 19th acm conference on computer-supported
  cooperative work \& social computing\/} (2016), ACM, pp.~418--431.

\bibitem{rae2017robotic}
{\sc Rae, I., and Neustaedter, C.}
\newblock Robotic telepresence at scale.
\newblock In {\em Proceedings of the 2017 CHI Conference on Human Factors in
  Computing Systems\/} (2017), ACM, pp.~313--324.

\bibitem{rascon2017ssl}
{\sc Rascon, C., and Meza, I.}
\newblock Localization of sound sources in robotics: A review.
\newblock {\em Robotics and Autonomous Systems 96\/} (2017), 184--210.

\bibitem{rascon2015hri}
{\sc Rascon, C., Meza, I., Fuentes, G., Salinas, L., and Pineda, L.~A.}
\newblock Integration of the multi-doa estimation functionality to human-robot
  interaction.
\newblock {\em International Journal of Advanced Robotic Systems 12}, 2 (2015),
  8.

\bibitem{rosebrock}
{\sc Rosebrock, A.}
\newblock Double 2 – features.
\newblock \url{https://www.doublerobotics.com/double2.html}.

\bibitem{rosebrock2017detect}
{\sc Rosebrock, A.}
\newblock Detect eyes, nose, lips, and jaw with dlib, opencv, and python.
\newblock
  \url{https://www.pyimagesearch.com/2017/04/10/detect-eyes-nose-lips-jaw-dlib-opencv-python/},
  April 10 2017.
\newblock dlib, Faces, Facial Landmarks, Libraries, Tutorials.

\bibitem{schmidt1986multiple}
{\sc Schmidt, R.}
\newblock Multiple emitter location and signal parameter estimation.
\newblock {\em IEEE transactions on antennas and propagation 34}, 3 (1986),
  276--280.

\bibitem{stoll2018wait}
{\sc Stoll, B., Reig, S., He, L., Kaplan, I., Jung, M.~F., and Fussell, S.~R.}
\newblock Wait, can you move the robot?: Examining telepresence robot use in
  collaborative teams.
\newblock In {\em Proceedings of the 2018 ACM/IEEE International Conference on
  Human-Robot Interaction\/} (2018), ACM, pp.~14--22.

\bibitem{tiwari}
{\sc Tiwari, S.}
\newblock Face detection in python using a webcam.
\newblock
  \url{https://realpython.com/face-detection-in-python-using-a-webcam/}.
\newblock © 2012–2019 Real Python.

\bibitem{vazquez2017autonomy}
{\sc V{\'a}zquez, M., Carter, E.~J., McDorman, B., Forlizzi, J., Steinfeld, A.,
  and Hudson, S.~E.}
\newblock Towards robot autonomy in group conversations: Understanding the
  effects of body orientation and gaze.
\newblock In {\em Proceedings of the 2017 ACM/IEEE International Conference on
  Human-Robot Interaction\/} (2017), ACM, pp.~42--52.

\bibitem{vazquez2016focus}
{\sc V{\'a}zquez, M., Steinfeld, A., and Hudson, S.~E.}
\newblock Maintaining awareness of the focus of attention of a conversation: A
  robot-centric reinforcement learning approach.
\newblock In {\em 2016 25th IEEE International Symposium on Robot and Human
  Interactive Communication (RO-MAN)\/} (2016), IEEE, pp.~36--43.

\bibitem{venolia2010embodied}
{\sc Venolia, G., Tang, J., Cervantes, R., Bly, S., Robertson, G., Lee, B., and
  Inkpen, K.}
\newblock Embodied social proxy: mediating interpersonal connection in
  hub-and-satellite teams.
\newblock In {\em Proceedings of the SIGCHI Conference on Human Factors in
  Computing Systems\/} (2010), ACM, pp.~1049--1058.

\end{thebibliography}









\end{document}